\documentclass[letterpaper, 10 pt, conference]{ieeeconf}  % Comment this line out if you need a4paper

\IEEEoverridecommandlockouts                              % This command is only needed if 
\usepackage{graphicx}
\graphicspath{ {./figures/} }
\usepackage[font=small]{subcaption}
\usepackage{xcolor}
\usepackage{algorithmic}
\usepackage{amsmath} % assumes amsmath package installed
\usepackage{amssymb}  % assumes amsmath package installed
\usepackage{float}

\newcommand{\G}{\mathcal G}
\newcommand{\x}{\mathcal X}
\newcommand{\xfree}{\mathcal X_{\text{free}}}
\newcommand{\xobs}{\mathcal X_{\text{obs}}}
\newcommand{\xgoal}{\mathcal X_{\text{goal}}}
\newcommand{\xinit}{x_{\text{init}}}

\newtheorem{problem}{Problem}

\usepackage{hyperref}
\hypersetup{
    colorlinks=true,
    linkcolor=black,
    filecolor=magenta,      
    urlcolor=blue,
}
\title{\LARGE \bf
Learned Critical Probabilistic Roadmaps for Robotic Motion Planning
}

\author{Brian Ichter$^{1}$, Edward Schmerling$^{2}$, Tsang-Wei Edward Lee$^{1}$, and Aleksandra Faust$^{1}$% <-this % stops a space
\thanks{$^{1}$Brian Ichter, Tsang-Wei Edward Lee, and Aleksandra Faust are with Robotics at Google, Mountain View, CA, USA
        {\tt\small $\{$ichter,tsangwei,faust$\}$@google.com}}%
\thanks{$^{2}$Edward Schmerling is with Waymo Research, Mountain View, CA, USA
        {\tt\small schmrlng@waymo.com}}%
}

\begin{document}

\maketitle
\thispagestyle{empty}
\pagestyle{empty}

%%%%%%%%%%%%%%%%%%%%%%%%%%%%%%%%%%%%%%%%%%%%%%%%%%%%%%%%%%%%%%%%%%%%%%%%%%%%%%%%
\begin{abstract}
Sampling-based motion planning techniques have emerged as an efficient algorithmic paradigm for solving complex motion planning problems. 
These approaches use a set of probing samples to construct an implicit graph representation of the robot's state space, allowing arbitrarily accurate representations as the number of samples increases to infinity. 
In practice, however, solution trajectories only rely on a few critical states, often defined by structure in the state space (e.g., doorways). 
In this work we propose a general method to identify these critical states via graph-theoretic techniques (betweenness centrality) and learn to predict criticality from only local environment features.
These states are then leveraged more heavily via global connections within a hierarchical graph, termed Critical Probabilistic Roadmaps. 
Critical PRMs are demonstrated to achieve up to three orders of magnitude improvement over uniform sampling, while preserving the guarantees and complexity of sampling-based motion planning.
A video is available at \url{https://youtu.be/AYoD-pGd9ms}.
\end{abstract}

%%%%%%%%%%%%%%%%%%%%%%%%%%%%%%%%%%%%%%%%%%%%%%%%%%%%%%%%%%%%%%%%%%%%%%%%%%%%%%%%

\section{Introduction}

Robot motion planning computes a collision free, dynamically feasible, and low-cost trajectory from an initial state to goal region \cite{lavalle2006planning}.
Sampling-based motion planning (SBMP) approaches, such as probabilistic roadmaps (PRMs) \cite{kavraki1994probabilistic} and rapidly-exploring random trees \cite{lavalle2000rapidly}, efficiently solve complex planning problems through using a set of probing samples to construct an implicit graph representation of the robot's state space.
To connect the initial state and goal region, PRMs search this roadmap graph and identify a sequence of states and local connections which the robot may traverse.
Though these algorithms can form arbitrarily accurate representations as the number of samples increases to infinity, in practice, only a few critical states are necessary to parameterize solution trajectories.
Often these critical states enjoy significant structure, e.g., entries to narrow passages, yet are only identified through exhaustive sampling \cite{ichter2018learning}.
Furthermore, when these states are identified~\cite{hsu2005hybrid,ichter2018learning}, traditionally they are treated with equal importance as less critical samples, e.g., samples in open regions.

We present a method that learns to recognize critical states and use them to construct a hierarchical PRM.
These critical states are quantified through \textit{betweenness centrality} \cite{freeman1977set}, a graph-theoretic measure of centrality based on a sample's importance to shortest paths through a graph, followed by a smoothing step that retains critical samples that are necessary for planning. 
Given this set of critical states, a neural network is learned to predict criticality from local, environmental features; this local focus enables scaling to complex environments.
Online, with a new, previously unseen planning problem, we construct a \textit{Critical PRM}, which samples a small number of critical states and a large number of non-critical states. 
The non-critical samples are connected locally, preserving the asymptotic optimality and complexity of SBMP.
The connection strategy for the critical states, however, is modified by connecting critical states to all samples, providing critical edges through the graph. 
The results in this work demonstrate the algorithm's generality and show a significant reduction in computation time to achieve the same success rate and cost as baselines.

\textit{Related Work.} 
Since the early days of sampling-based motion planning \cite{kavraki1994probabilistic}, researchers have sought to develop improved sampling techniques that bias samples towards important regions of the state space \cite{hsu1998finding, boor1999gaussian} or cover the state space more efficiently \cite{branicky2001quasi}. 
Several works have considered deterministic sequences for covering spaces evenly \cite{branicky2001quasi,lavalle2004relationship,janson2018deterministic}.
To bias samples, several previous works have used heuristically driven approaches to sample more frequently near obstacles \cite{boor1999gaussian}, near narrow passages \cite{hsu1998finding}, or to adaptively combine several of such approaches \cite{hsu2005hybrid}.
Others have used workspace decompositions to identify regions of interest \cite{kurniawati2004workspace, van2005using}. 
Another promising approach~\cite{gammell2015batch} adaptively samples only in regions that may improve the current solution.
These methods are generally limited in their ability to sample the state space or consider local planner policies.

One recent approach has been to learn sample distributions to directly sample from in sampling-based motion planning \cite{zucker2008adaptive, ichter2018learning,zhang2018learning,chamzas2019using,molina2019identifying}.
\cite{ichter2018learning} and \cite{zhang2018learning} use offline solution trajectories to learn a distribution of samples and bias sampling towards regions where an optimal solution might lie.
\cite{chamzas2019using} learns a local library sampler to bias sampling towards regions likely to contain the solution.
\cite{molina2019identifying} learns to identify critical regions for sampling based on images of successful solutions and leverages these samples by growing trees from them.
\cite{kumar2019lego} proposes several heuristic approaches for identifying diverse bottleneck states to plan with sparse roadmaps.
In this work, we identify critical regions from graph theoretical techniques in the state space and incorporate these samples into a hierarchical Critical Roadmap, which allows the critical samples to connect throughout the state space.
Furthermore, most previous work has focused on a single-query setting, where the initial and goal state can be heavily leveraged in sample selection and where complex, large environments can be difficult to learn from. 
Instead, Critical PRMs are targeted towards multi-query settings and only require local information to choose samples.

Finally, we note the connections between Critical PRM and other areas of machine learning.
Within Reinforcement Learning (RL), the exploration problem is not unlike the motion planning problem we consider herein.
In this context, critical states may be used to identify and subsequently reward useful subgoals, allowing more efficient exploration \cite{wu2018laplacian,goyal2019infobot}.
These critical states can further be used to identify discrete skills to enable learned hierarchies \cite{bacon2013bottleneck, csimcsek2009skill, solway2014optimal}.
Within planning and control, recent work has sought to learn low-dimensional representations of important regions of the state space \cite{watter2015embed,banijamali2017robust,ichter2019robot}.

\textit{Statement of Contributions.}
The contributions of this paper are threefold.
First, we present a methodology for identifying and subsequently learning to identify critical states for the optimal motion planning problem.
The criticality of a node is based on the graph-theoretic betweenness centrality, which allows states to be identified in problems with complex environments, state spaces, and local planners.
The learned regressor further can identify criticality from only local features, allowing scaling to more complex environments and requiring less training data.
Second, we present an algorithm for leveraging these critical states as more important through a hierarchical PRM, termed the Critical PRM.
In particular, these states are globally connected through the state space, along with a set of locally connected uniform states. 
This allows critical states and connected edges to serve as highways through the state space.
Third, we demonstrate the Critical PRM algorithm on a number of motion planning problems as well as compare to state of the art baselines.
Our findings show that Critical PRM can outperform other methods by up to three orders of magnitude in computation time to a success rate and one in cost.
Furthermore, we show that Critical PRM can scale to the complexity of real-world planning problems.

\textit{Organization.}
In Section \ref{sec:problem}, we introduce the optimal motion planning problem approached herein.
In Section \ref{sec:identify}, we overview the process of learning to identify critical samples.
In Section \ref{sec:cprm}, we describe the Critical PRM algorithm.
In Section \ref{sec:exp}, we demonstrate the generality and speed improvements of Critical PRMs as well as show the performance on a real robot.
Finally, in Section \ref{sec:conclusions}, we overview conclusions and outline future directions.

\section{Problem Statement}\label{sec:problem}

The goal of this work is to efficiently solve optimal motion planning problems by learning to identify and effectively leveraging states critical to their solutions.
Informally, solving the optimal motion planning problem entails finding the shortest free path from an initial state to a goal region, if one exists.
For complex problems, there exist formulations that include kinematic, differential, or other more complex constraints \cite{schmerling2015optimal,lavalle2006planning}. 
The geometric motion planning problem, a simple version of the problem, is defined as follows.
Let $\x = [0,1]^d$ be the state space, with $d\in \mathbb{N},d\geq 2$.
Let $\xobs$ denote the obstacle space, $\xfree = \x \setminus \xobs$ the free state space, $\xinit\in\xfree$ the initial condition, and $\xgoal \subset \xfree$ the goal region.
A path is defined by a continuous function, $s: [0,1] \to \mathbb{R}^d$.
We refer to a path as \emph{collision-free} if $s(\tau) \in \xfree$ for all $\tau \in [0,1]$, and \emph{feasible} if it is collision-free, $s(0) = \xinit$, and $s(1) \in \xgoal$. 
We thus wish to solve, 

\vspace{0.15cm}
\begin{problem}[Optimal motion planning]
Given a motion planning problem $(\xfree, \xinit, \xgoal)$ and a cost measure $c$, find a feasible path $s^\ast$ such that $c(s^\ast) = \min \{c(s):s$ is feasible$\}$. If no such path exists, report failure.
\end{problem}
\vspace{0.15cm}

Even simple forms of the motion planning problem are known to be PSPACE-complete \cite{lavalle2006planning}, and thus one often turns to approximate methods to solve the problem efficiently.
In particular, sampling-based motion planning techniques have emerged as one such state of the art approach. 
These algorithms avoid explicitly constructing the problem's state space and instead build an approximate, implicit representation of the state space.
This representation is constructed through a set of probing samples, each a potential state the robot may be in.
A graph (or tree) is then built by connecting these samples to their local neighbors via a local planner under the supervision of a black-box collision checker \cite{lavalle2006planning}.
Finally, given an initial and goal state, this representation can be searched to find a trajectory connecting the two.
In this work we focus on a multi-query setting, focusing on identifying critical states and connecting critical states globally.

\section{Critical Sample Identification and Learning}\label{sec:identify}

\subsection{Identifying Critical States} 

The first question we seek to answer is what defines a critical sample for the optimal motion planning problem.
If we consider a human navigating an indoor environment, these critical states may be doorways and non-critical states may be hallways or other open regions. 
In the context of geometric motion planning discovering these narrow passages is often the bottleneck \cite{hsu1998finding}.
However, as problems increase in difficulty, either due to more complex environments (cluttered and unstructured) or more complex robotic systems (differential constraints, rotational DoFs, complex local policies, stochasticity), it is not clear how such concepts can be utilized.
We thus wish to devise a principled, general method for extracting and learning to identify critical states.

\begin{figure}[t]
    \centering
    \begin{subfigure}[b]{0.21\textwidth}
        \includegraphics[width=\textwidth]{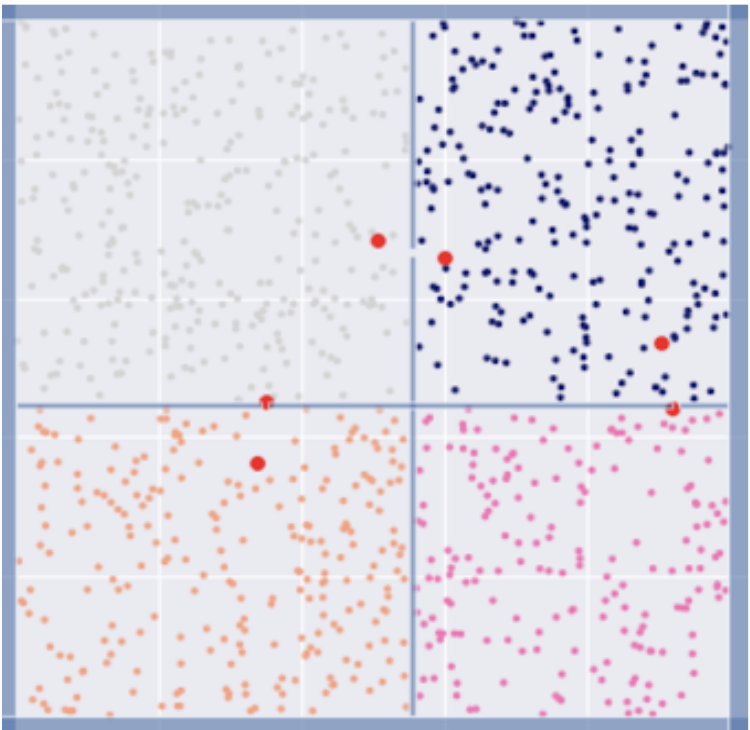}
        \caption{Label Propagation}
        \label{fig:lpa}
    \end{subfigure}
    \begin{subfigure}[b]{0.21\textwidth}
        \includegraphics[width=\textwidth]{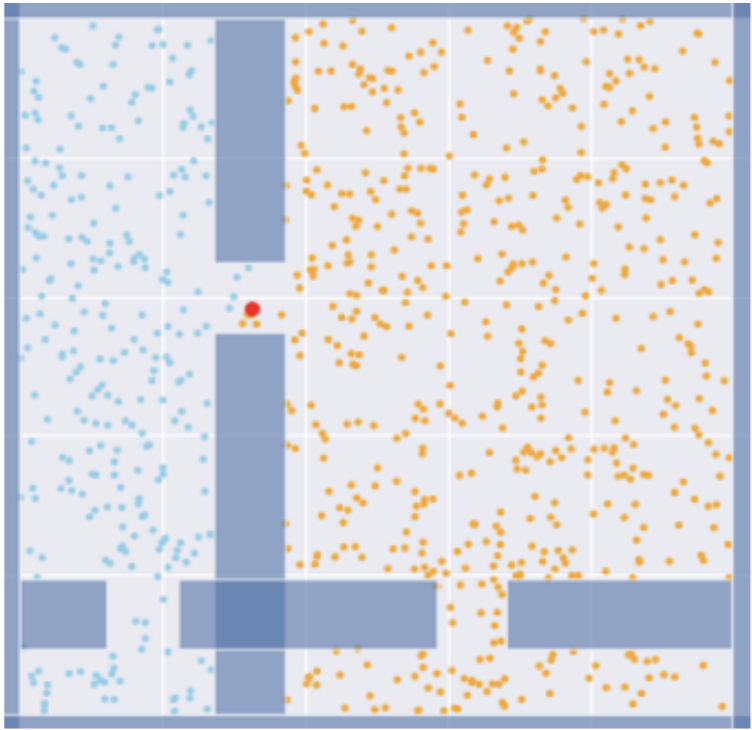}
        \caption{Label Propagation}
        \label{fig:lpa_fail}
    \end{subfigure}
    \begin{subfigure}[b]{0.21\textwidth}
        \includegraphics[width=\textwidth]{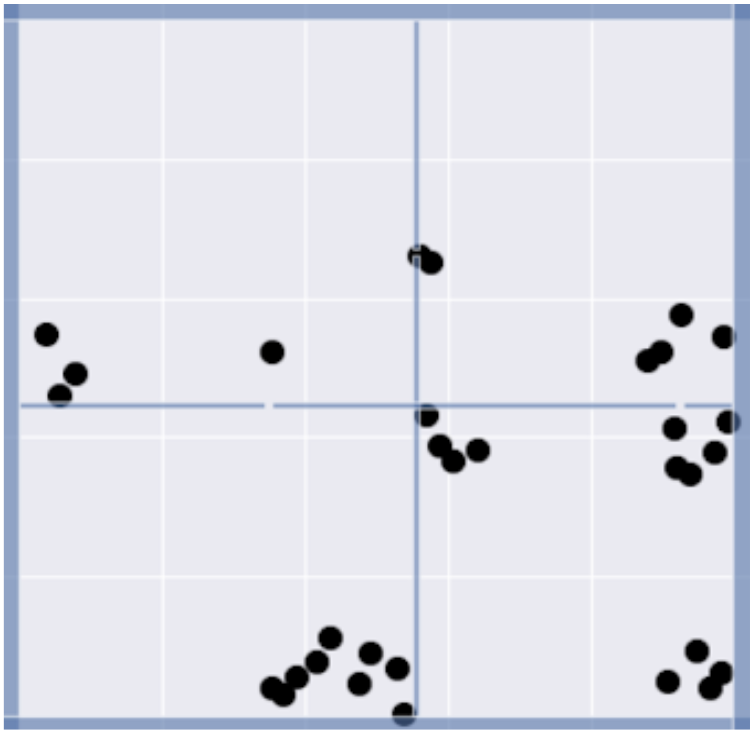}
        \caption{Minimum k-cuts}
        \label{fig:k_cuts}
    \end{subfigure}
    \begin{subfigure}[b]{0.21\textwidth}
        \includegraphics[width=\textwidth]{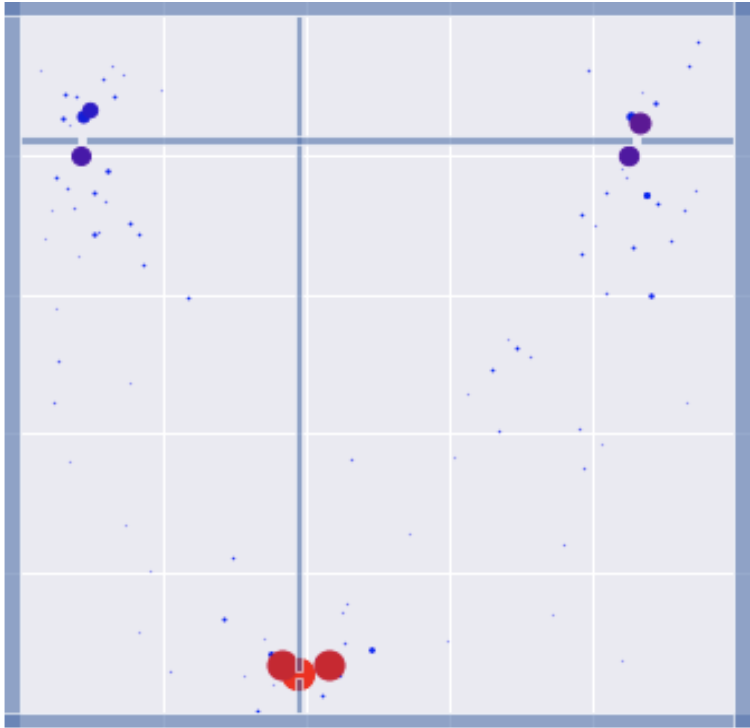}
        \caption{Betweenness}
        \label{fig:sample_data}
    \end{subfigure}
    \caption{
    Approaches for identification of critical samples: 
    (\ref{fig:lpa}-\ref{fig:lpa_fail}) Label propagating approaches fail with moderately narrow passages.
    (\ref{fig:k_cuts}) Minimum k-cuts identify several non-critical samples.
    (\ref{fig:sample_data}) Betweenness centrality was found to be principled and perform well. The size and color is proportional to criticality.}    
    \label{fig:prms}
\end{figure}

Several approaches exist to compute a sample's ``criticality'' in the state space; the most promising we considered were: label propagation \cite{raghavan2007near}, minimum k-cuts \cite{goldschmidt1994polynomial}, and betweenness centrality \cite{freeman1977set}.
Label propagation algorithms seek to break graphs into communities, where the transition between communities may be considered a bottleneck state.
Label propagation finds reasonable results for very narrow passage problems, but the results are unstable and poorly defined for problems with less constrained bottlenecks (Fig \ref{fig:lpa_fail}).
Minimum $k$-cut algorithms seek to find the minimum-weighted cuts in a graph partition the graph into $k$ components.
Minimum $k$-cuts can identify many critical samples, but require a fixed number of cuts be provided which can result in too few cuts, thus ignoring critical regions, or too many, thus identifying non-critical regions, e.g., corners, Fig. \ref{fig:k_cuts}.
Furthermore, minimum $k$-cuts is significantly slower than the other approaches.  
Ultimately, we selected \textit{betweenness centrality}, a graph-theoretic measure of the importance of each node to shortest paths through a graph. 

\begin{figure}[t]
    \centering
    \begin{subfigure}[b]{0.22\textwidth}
        \includegraphics[width=\textwidth]{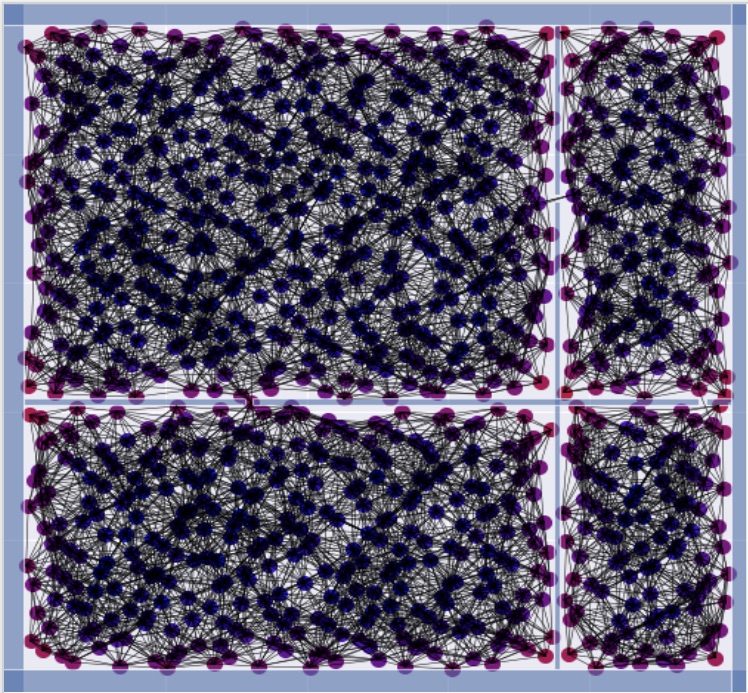}
        \caption{PRM}
        \label{fig:prm_build}
    \end{subfigure}
    \begin{subfigure}[b]{0.225\textwidth}
        \includegraphics[width=\textwidth]{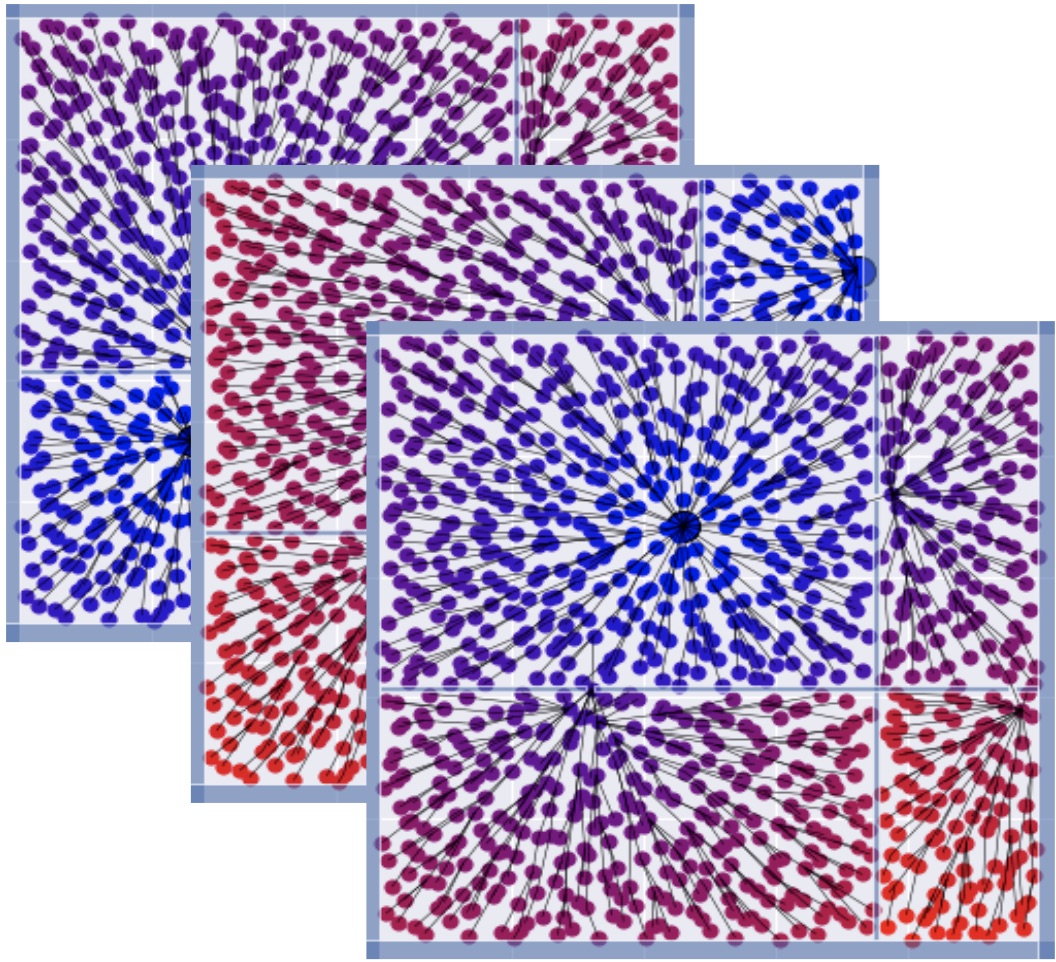}
        \caption{Solve Problems}
        \label{fig:prm_solves}
    \end{subfigure}
    \begin{subfigure}[b]{0.22\textwidth}
        \includegraphics[width=\textwidth]{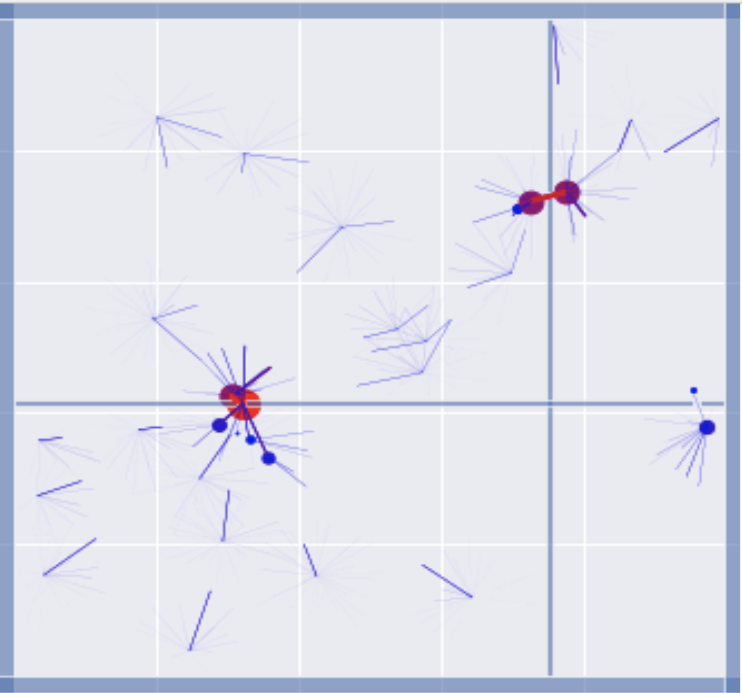}
        \caption{Criticality Data}
        \label{fig:identify_criticality}
    \end{subfigure}
    \begin{subfigure}[b]{0.22\textwidth}
        \includegraphics[width=\textwidth]{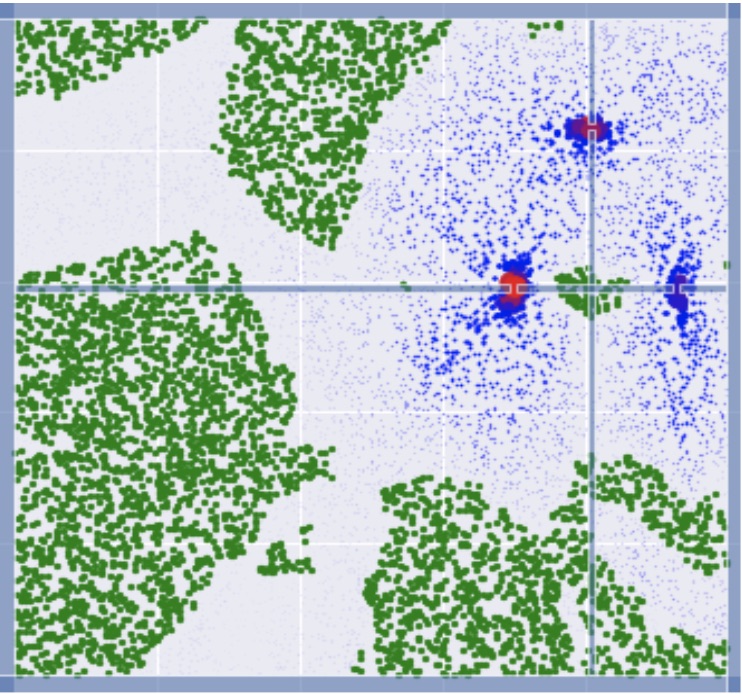}
        \caption{Learned Criticality}
        \label{fig:learn_criticality}
    \end{subfigure}
    \caption{
    The critical sample identification and learning process:
    (\ref{fig:prm_build}) Build PRMs on a family of training environments.
    (\ref{fig:prm_solves}) Solve several one to all planning problems.
    (\ref{fig:identify_criticality}) Identify points via betweenness centrality. The size (larger more critical) and color (red more critical, blue less) of each sample is proportional to criticality, the lines colors indicate how often an edge is used and are only for visualization.
    (\ref{fig:learn_criticality}) In a new environment, the criticality prediction network predicts the criticality of each sample (green indicates not critical, blue to red is colored in increasing criticality). Ultimately, Critical PRM critical states are sampled proportional to their criticality.}
    \label{fig:criticality}
\end{figure}

As outlined in Fig. \ref{fig:criticality}, betweenness centrality is computed by counting the number of all-pairs shortest paths that pass through a specific node.
We make two alterations to betweenness centrality to adapt it to the motion planning problem and the complexity of PRMs.
First, we only compute an approximate value by solving $m$ shortest path problems with a randomly chosen initial node (sampled without replacement) to all other graph nodes (note if $m = n$, this is exact).
Each time a node is used in a shortest path, its centrality score is incremented.
Secondly, we add a smoothing step to discount samples that can be skipped along the shortest path. 
Essentially, for a collision-free path that traverses nodes $x_i, x_{i+1}, x_{i+2}$, if the connection between node $x_i$ and $x_{i+2}$ is collision-free, then node $x_{i+1}$ is not critical to the path, and thus its score should not be incremented.
This step is necessary to eliminate samples that are simply in the free space trajectory between critical samples and used due to the limited $r_n$ connection radius.
Crucially, it also allows critical states to be identified in part by local environment features, enabling more compact environment representations and increased data efficiency, and thus better scaling to complex environments.

\subsection{Learning to Recognize States}

The first phase of the algorithm learns to identify critical samples from a set of PRMs generated for a family of training environments. 
This phase generates the critical sample dataset and then trains a predictor deep neural net model conditioned on the planning environment to output the criticality of a sample.
This methodology allows the neural network to generalize to new environments at test time.
Furthermore, due to the smoothing step, which discounts the criticality of states if they can be skipped in the local region, this network only needs a local representation of the environment.
This allows for much more scalable training in complex problems, as demonstrated in Section \ref{sec:exp}.
Note that here we learn to predict sample criticality and sample accordingly rather than learning a distribution of states. 
Though the distributional approach may allow more precise critical regions to be learned, this regressor approach only requires local features and may handle the multi-modality of critical states more effectively.

\textit{Dataset Creation.} For a given set of state spaces $\xfree$ in a training set, we construct a standard PRM $\G$ with samples $\{x_i\}_{i \in [1..n]} \in \xfree$ and edges $(x_i, x_j)$ for $i,j \in [1..n]$ if and only if the trajectory from sample $x_i$ to $x_j$ is collision free and the samples are within a connection radius $r_n$, as defined in \cite{karaman2011sampling}.
With these roadmaps in hand, we compute the criticality of each sample via the betweenness centrality as described above.

\textit{Training.} The computed centrality values become labels for training a neural network $h_\theta(x,y(x)),$ parameterized by $\theta,$ where $x \in \xfree$ is a state sample and $y(x)$ is a representation of $\xfree$. Herein, $y(x)$ is a representation of the local environment around $x$, e.g., an occupancy grid. We minimize $L_2$ loss to learn $\theta$.

\begin{figure*}
\noindent\begin{minipage}{\textwidth}\centering
\begin{minipage}{.5\textwidth}
  \captionsetup{singlelinecheck = false, format= hang, justification=raggedright, font=small, labelsep=space}
    \caption{{Online Critical PRM Construction}} 
    \begin{algorithmic}[1]\label{alg:cprm}\small
        \REQUIRE Planning problem $(\xfree, \xinit, \xgoal)$, $\lambda$, $\Gamma$, $n$
        \STATE Construct conditioning variables $y(x_i)$ (e.g., occupancy grid). \label{line:cond}
        \STATE Sample $\Gamma n$ states and compute criticality with $h_\theta(x_i,y(x_i))$. \label{line:sample}
        \STATE Select $\lambda \log(n)$ critical samples proportional to criticality. \label{line:sample_crit}
        \STATE Select $n - \lambda \log(n)$ uniform samples. \label{line:sample_uniform}
        \STATE Connect non-critical samples within an $r_n$ radius \cite{karaman2011sampling}.\label{line:connect_uniform}
        \STATE Connect critical samples to all samples (see footnote 1).\label{line:connect_crit}
        \STATE Connect $\xinit$ and $\xgoal$ globally into the Critical PRM.\label{line:init_goal}
        \STATE Search Critical PRM for shortest path from $\xinit$ to $\xgoal$.\label{line:search}
    \end{algorithmic}
\end{minipage}
\begin{minipage}{.03\textwidth}
    $\quad$
\end{minipage}
\begin{minipage}{.22\textwidth}
    \includegraphics[width=\textwidth]{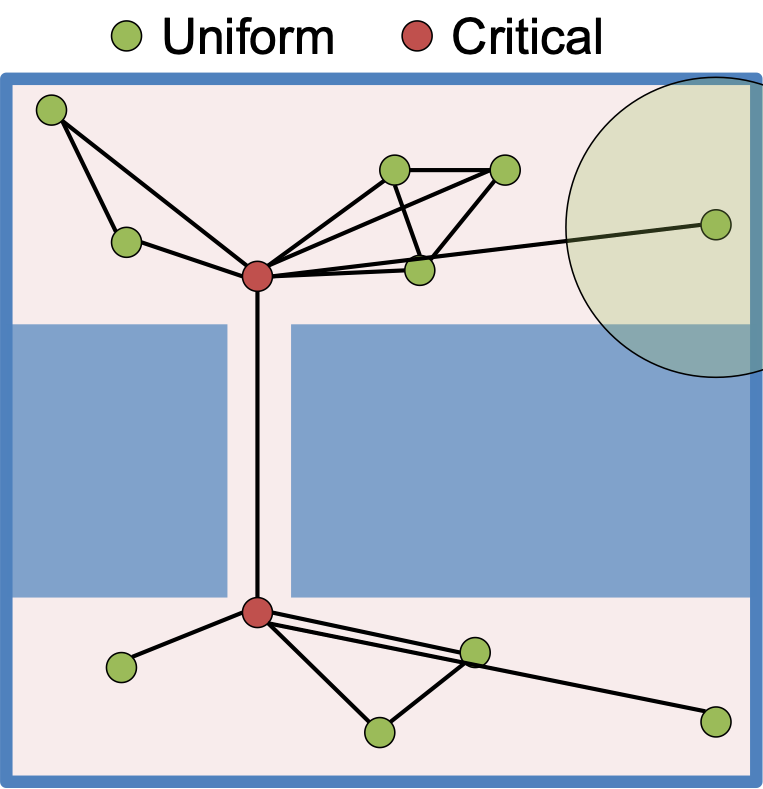}
\end{minipage}
\end{minipage}
\end{figure*}

\section{Critical Probabilistic Roadmaps}\label{sec:cprm}

% \begin{figure*}
% % \noindent\begin{minipage}{\textwidth}\centering
% % \begin{minipage}{.5\textwidth}
%     % \algsetup{linenodelimiter=}
% %   \vspace{0.4cm}
%   \begin{algorithmic}
%   \label{alg:cprm}\small
%     % \caption{\textbf{Online Critical PRM Construction}}
%      \textbf{Input:} Planning problem $(\xfree, \xinit, \xgoal)$, $\lambda$, $\Gamma$, $n$ \label{line:input}\\
%      Construct conditioning variables $y(x_i)$ (e.g., occupancy grid). \label{line:cond}\\
%      Sample $\Gamma n$ states and compute criticality with $h_\theta(x_i,y(x_i))$. \label{line:sample}\\
%      Select $\lambda \log(n)$ critical samples proportional to criticality. \label{line:sample_crit}\\
%      Select $n - \lambda \log(n)$ uniform samples. \label{line:sample_uniform}\\
%      Connect non-critical samples within an $r_n$ radius \cite{karaman2011sampling}.\label{line:connect_uniform}\\
%      Connect critical samples to all samples (see footnote 1).\label{line:connect_crit}\\
%      Connect $\xinit$ and $\xgoal$ globally into the Critical PRM.\label{line:init_goal}\\
%      Search Critical PRM for shortest path from $\xinit$ to $\xgoal$.\label{line:search}\\
%   \end{algorithmic}
% % \end{minipage}
% % \begin{minipage}{.03\textwidth}
% %   $\quad$
% %   \end{minipage}
% \begin{minipage}{.22\textwidth}
%   \includegraphics[width=\textwidth]{critPRM}
% \end{minipage}
% % \end{minipage}
% \end{figure*}

Given the ability to identify critical samples in a previously unseen environment, the next key question is how to best leverage these important samples.
Previous works have general sampled them at a higher rate, but considered them of equal importance beyond that \cite{hsu2005hybrid,ichter2018learning}.
Herein, we propose a hierarchical graph, called a Critical Probabilistic Roadmap (Critical PRM), that globally connects critical states, along with a bed of locally connected uniform states.
This allows these critical states to act as primary hubs within the space, while preserving the theoretical guarantees of sampling-based motion planning via the uniform samples.

Given a new planning environment, the online portion of the Critical PRM algorithm proceeds as follows (and outlined in Fig. \ref{alg:cprm}.
The value of $\Gamma$ allows the critical samples to be chosen from a more dense covering of the state space.
Given a planning problem in a new environment with free space $\xfree$, corresponding environmental input $y$, a sample budget of $n,$ and a constant $\lambda$ that controls the number of critical samples, we first select $\Gamma n$ samples $\{x_i\}_{i \in [1..\Gamma n]} \in \xfree$ and predict their criticality with $h_\theta(x_i,y(x_i))$ (Line \ref{line:cond}-\ref{line:sample}). 
Next, to select $\lambda \log(n)$ critical samples, the samples are stochastically drawn with a probability proportional to their criticality  (Line \ref{line:sample_crit}). 
We refer to the remaining samples $n - \lambda \log(n)$ as non-critical samples  (Line \ref{line:sample_uniform}).
To connect the Critical PRM, we connect the non-critical samples locally, only with neighbors within an $r_n$ connection radius, as detailed in \cite{karaman2011sampling} (Line \ref{line:connect_uniform}). 
In contrast, the critical samples are connected globally, to all other samples regardless of distance (Line \ref{line:connect_crit}).\footnote[1]{For very large spaces or costly local planners, this connection radius can instead be a large constant value.}
Finally, given a planning problem with an initial and goal region, we connect them into the roadmap globally, and search the roadmap for the shortest connecting path (Line \ref{line:init_goal}-\ref{line:search}). 
Fig. \ref{fig:prm_critical_20} shows a Critical PRM example with 20 samples versus a uniformly sampled PRM with 1000 samples in Fig. \ref{fig:prm_uniform_1k}, wherein the Critical PRM achieves better connectivity than a standard PRM with 50x fewer samples.

\textit{Complexity:} The complexity of Critical PRM remains $\mathcal{O}(n \log(n))$ as the $n-\lambda \log(n)$ uniform samples are locally connected and maintain the standard $\mathcal{O}(n \log(n))$ complexity of PRM \cite{karaman2011sampling}. 
The $\lambda \log(n)$ critical samples are connected to all $n$ neighbors, requiring no nearest neighbor lookup and $n \lambda \log(n)$ constant time connections and collision checks.

\textit{Probabilistic Completeness and Asymptotic Optimality:} The theoretical guarantees of probabilistic completeness and asymptotic optimality from \cite{janson2015fast,janson2018deterministic,karaman2011sampling} hold for this method by adjusting any references to $n$ (the number of samples) to $(n-\lambda \log(n))$ (the number of uniform samples in our methodology).
This result is detailed in Appendix D of \cite{janson2015fast} and Section 5.3 of \cite{janson2018deterministic}, which show that adding samples can only improve the solution.

\section{Experiments}\label{sec:exp}

In this section we evaluate the Critical PRM algorithm on several motion planning problems and on robot.
The results in this work were implemented in a mix of Python and Julia \cite{bezanson2012julia} along with TensorFlow~\cite{abadi2016tensorflow}.
The network architectures are fully connected for 1D environment inputs, convolutional for 2D inputs, and 3D convolutional for 3D inputs.
The loss for each problem is a log mean squared error on a sample's criticality.
Each training dataset was composed of 50\% critical states (states with criticality greater than 0) and 50\% non-critical states.
Note that we make comparisons for several problems well-tuned to Hybrid sampling \cite{hsu2005hybrid}, however we do not make comparisons to previous learning-based approaches targeted towards single-query settings, as these heavily rely on initial and goal states and have difficulty scaling to the large complex environments that the local nature of Critical PRM can cope with.

\subsection{Narrow Passage Environment} \label{subsec:NPE}

\begin{figure*}[t]
    \centering
    $\quad$
    \begin{subfigure}[b]{0.18\textwidth}
        \includegraphics[width=\textwidth]{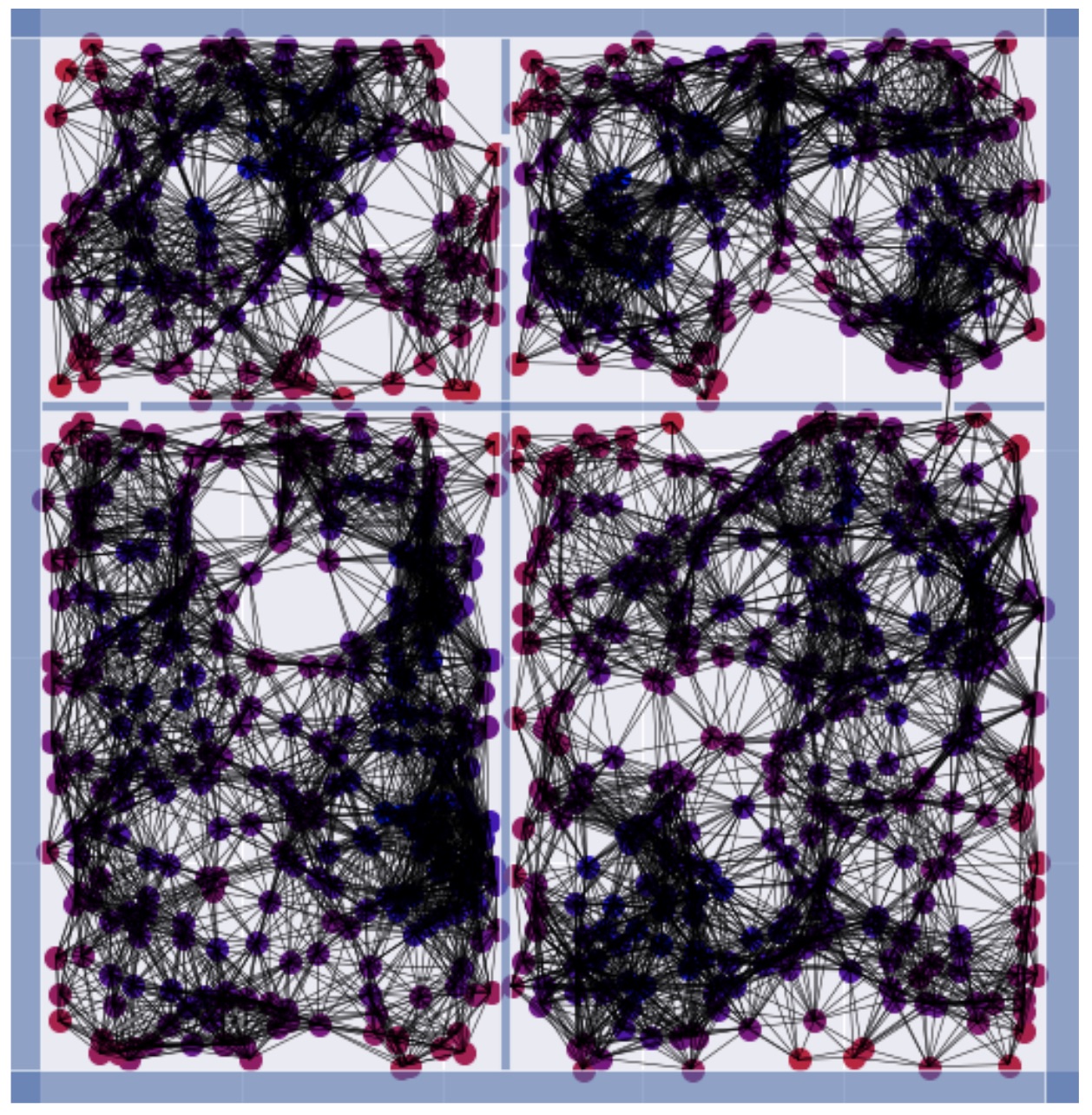}
        \caption{Uniform PRM}
        \label{fig:prm_uniform_1k}
    \end{subfigure}
    $\qquad$
    \begin{subfigure}[b]{0.18\textwidth}
        \includegraphics[width=\textwidth]{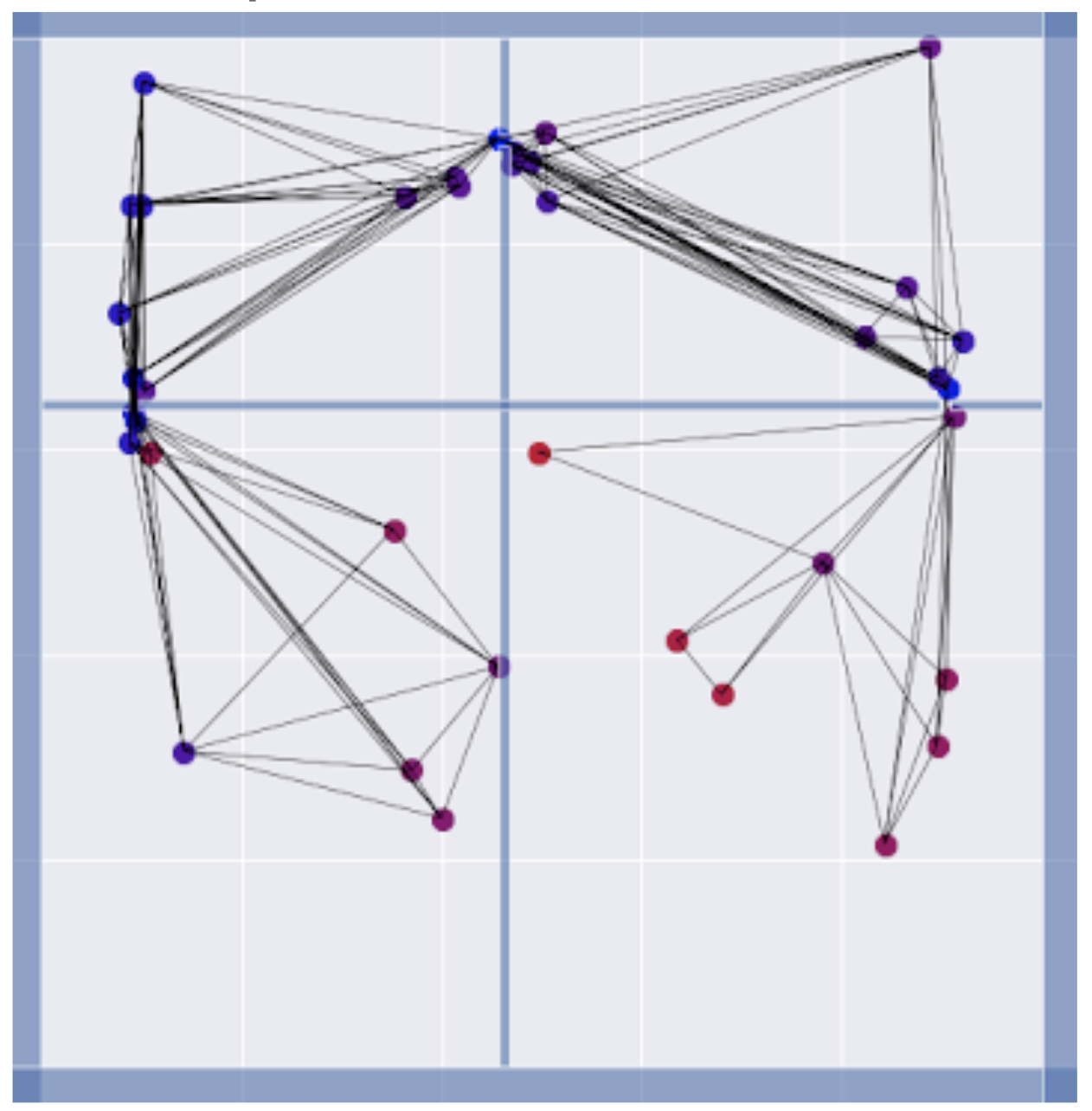}
        \caption{Critical PRM}
        \label{fig:prm_critical_20}
    \end{subfigure}
    \begin{subfigure}[b]{0.235\textwidth}
        \includegraphics[width=\textwidth]{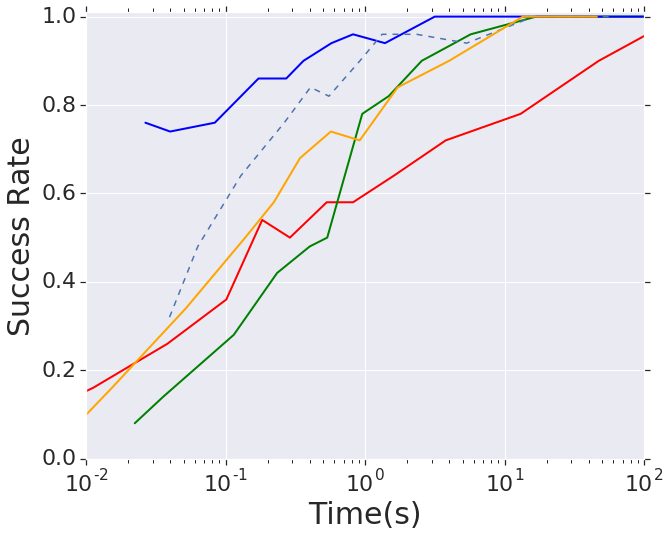}
        \caption{2D: Time (s) vs. Success}
        \label{fig:time_vs_success}
    \end{subfigure}
    \begin{subfigure}[b]{0.235\textwidth}
        \includegraphics[width=\textwidth]{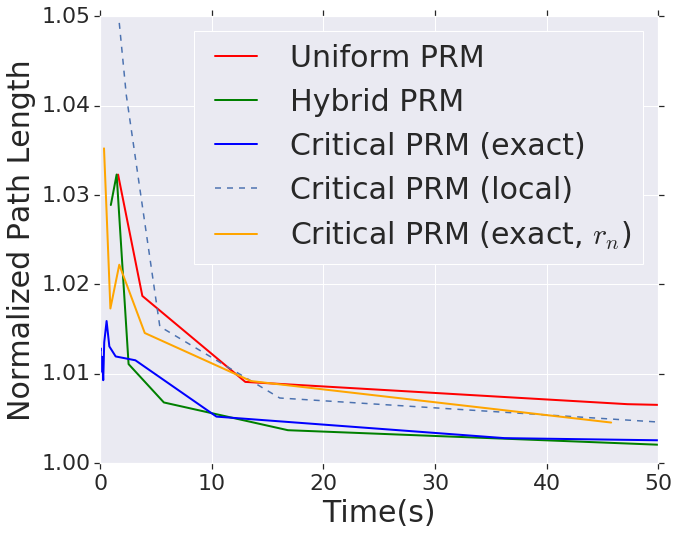}
        \caption{2D: Time (s) vs. Cost}
        \label{fig:time_vs_cost}
    \end{subfigure}
        \begin{subfigure}[b]{0.2\textwidth}
        \includegraphics[width=\textwidth]{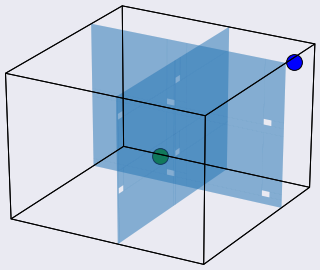}
        \caption{3D Problem}
        \label{fig:3d_problem_0}
    \end{subfigure}
    $\qquad$
    \begin{subfigure}[b]{0.163\textwidth}
        \includegraphics[width=\textwidth]{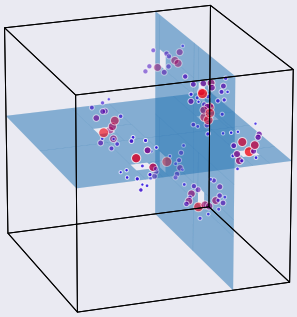}
        \caption{Learned Criticality}
        \label{fig:3d_problem_1}
    \end{subfigure}
    \begin{subfigure}[b]{0.235\textwidth}
        \includegraphics[width=\textwidth]{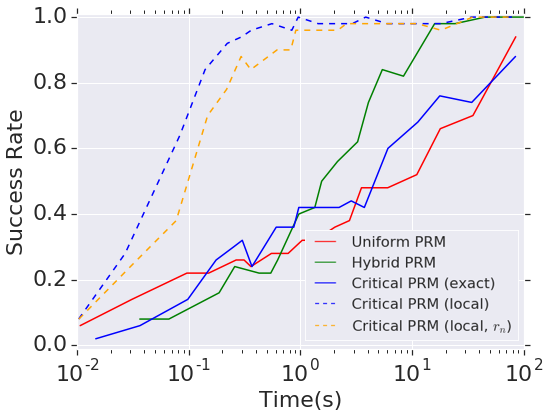}
        \caption{3D: Time (s) vs. Success}
        \label{fig:3d_time_vs_success}
    \end{subfigure}
    \begin{subfigure}[b]{0.235\textwidth}
        \includegraphics[width=\textwidth]{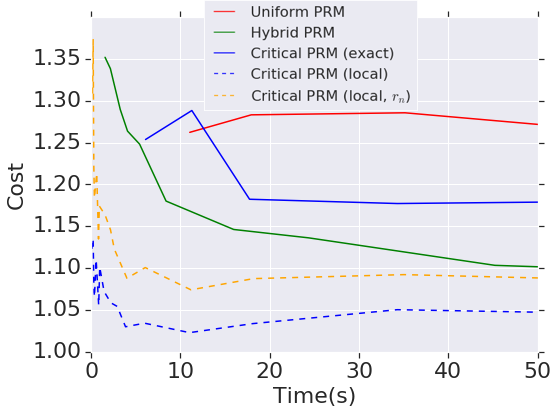}
        \caption{3D: Time (s) vs. Cost}
        \label{fig:3d_time_vs_cost}
    \end{subfigure}
    \caption{
    With 50x fewer samples, the Critical PRM (\ref{fig:prm_critical_20}) fully connects a space that uniform PRM (\ref{fig:prm_uniform_1k}) cannot. (\ref{fig:time_vs_success}) Success rate and (\ref{fig:time_vs_cost}) cost for Critical PRM (blue) and uniform PRM (red) shows orders of magnitude improvement (over 50 problems). Furthermore, the use of global connections for critical states is responsible for an order of magnitude improvement compared to critical states with a standard $r_n$ radius (orange). (\ref{fig:3d_problem_0}-\ref{fig:3d_time_vs_cost}) 3D problem demonstrates similar results and that Critical PRM with local features (a local occupancy grid) performs well in more complex environments.}
    \label{fig:prms}
\end{figure*}

As a proof of concept, we consider randomly generated 2D and 3D narrow passage environments shown in Figs. \ref{fig:prm_uniform_1k}-\ref{fig:time_vs_cost} and \ref{fig:3d_problem_0}-\ref{fig:3d_time_vs_cost}.
These environments provide both clear learned samples and allow comparison to previous heuristic methods tuned for narrow passages \cite{hsu2005hybrid}.
Two critical prediction networks were trained for each environment with exact and local workspace representations.
The first network is given an \textit{exact} representation of the workspace as input: the $(x,y)$-coordinates of each gap.
The second network is given a \textit{local} representation of the workspace as input. For 2D, the workspace is divided into a $100\times100$ occupancy grid, of which the local $10\times10$ occupancy grid is fed into the network. For 3D, the workspace is divided into $36\times36\times36$ and locally $12\times12\times12$.
2D used $\lambda = 2$ and 3D used $\lambda = 10$.
The training data consisted of 1k problems and 100k example states.
For comparisons, we also consider a standard, uniform PRM \cite{kavraki1994probabilistic} and Hybrid sampling PRM \cite{hsu2005hybrid}, which specifically biases samples towards obstacles and narrow passages.
For the network with better performance on each problem, we consider the effect of globally connecting critical samples by showing results for sampling critical states, but only $r_n$ local connections (instead with the connection radius $r_n$).

For each problem, Critical PRM outperforms both uniform and hybrid sampling in cost and success rate vs. computation time (s).
For 2D, the exact representation achieves better initial performance compared to the occupancy grid input, though the occupancy grid input performs similarly well at higher computation times.
However for 3D, due to the complexity of the problem, the local representation vastly outperformed the exact even with a significantly higher dimensionality.
This is because each sample's local environment is unique, allowing for 100k different environment training inputs instead of only 1k unique inputs for the exact.
Furthermore, while the Critical PRM with a standard connection radius outperforms uniform sampling, the global connections improve both cost and success rate substantially.

\subsection{Rigid Body Planning}\label{sec:rigid}

\begin{figure}[t]
    \centering
    \begin{subfigure}[b]{0.215\textwidth}
        \includegraphics[width=\textwidth]{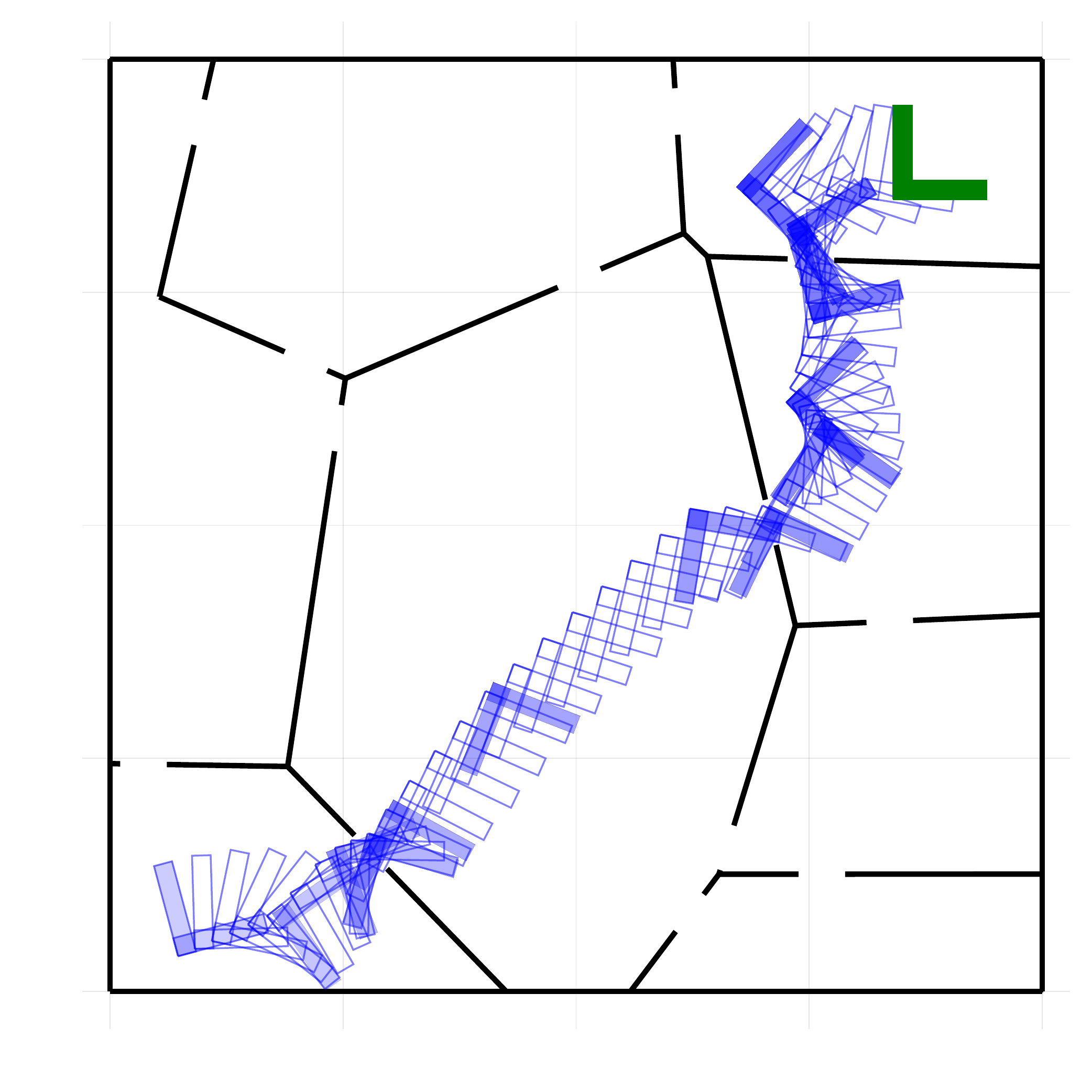}
        \caption{SE(2) Planning}
        \label{fig:SE2_solution}
    \end{subfigure}
    \begin{subfigure}[b]{0.21\textwidth}
        \includegraphics[width=\textwidth]{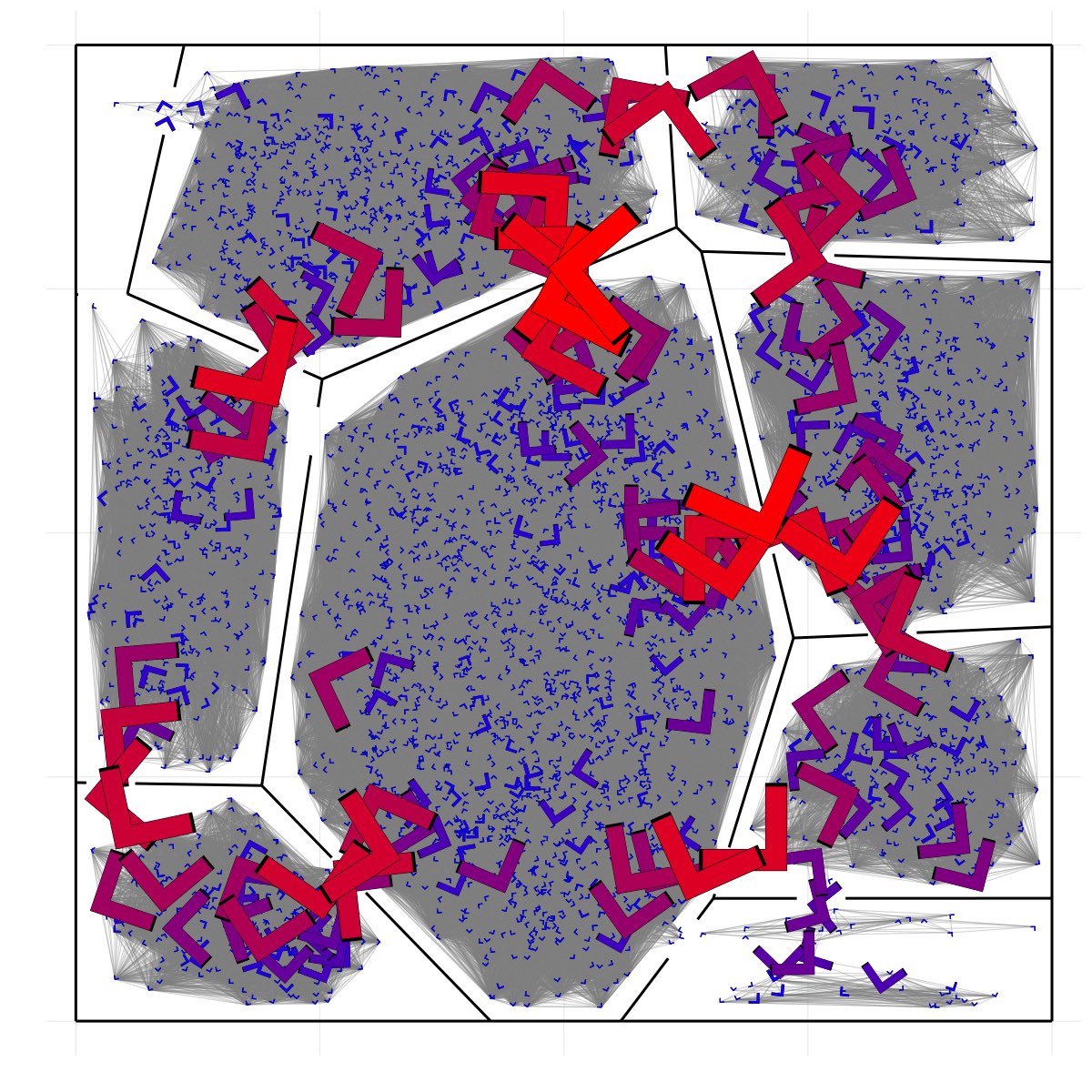}
        \caption{Ground Truth Criticality}
        \label{fig:SE2_ground_truth}
    \end{subfigure}
    \begin{subfigure}[b]{0.22\textwidth}
        \includegraphics[width=\textwidth]{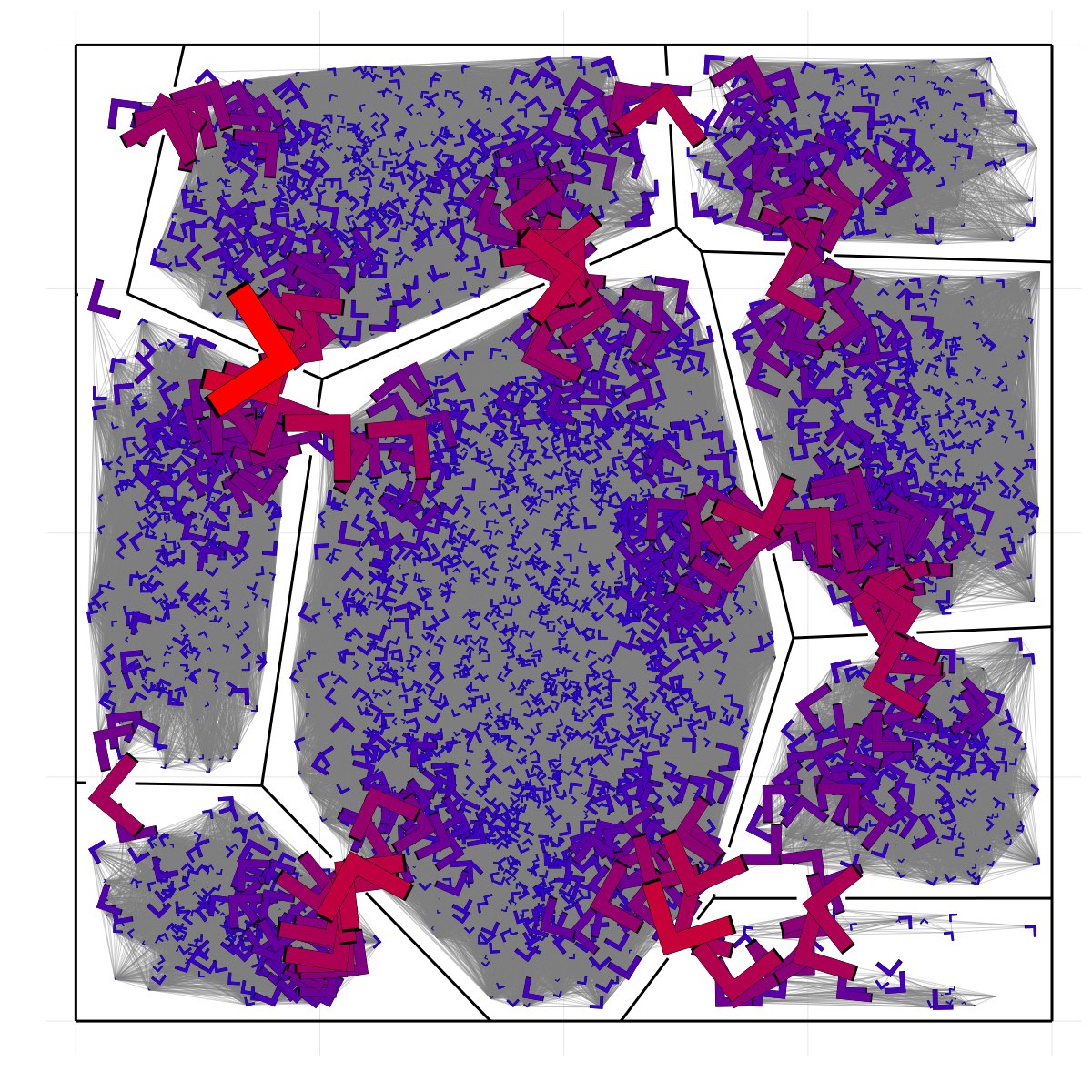}
        \caption{Learned Criticality}
        \label{fig:SE2_learned}
    \end{subfigure}
    \begin{subfigure}[b]{0.22\textwidth}
        \raisebox{0.13\textwidth}{
        \includegraphics[width=\textwidth]{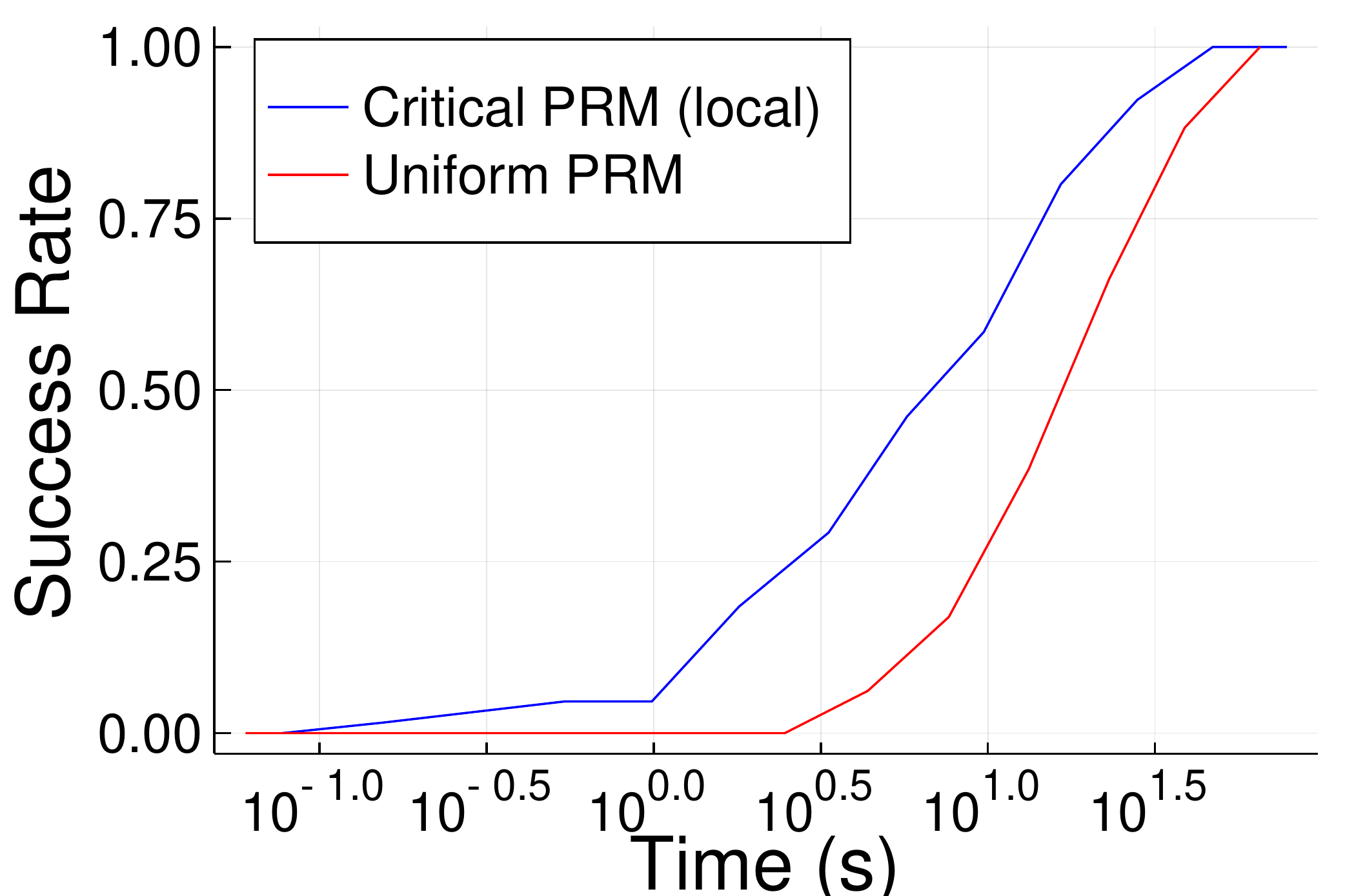}}
        \caption{Time (s) vs. Success}
        \label{fig:SE2_time_vs_success}
    \end{subfigure}
    \caption{ 
    Learning criticality for SE(2) rigid body motion planning (\ref{fig:SE2_solution}). The ground truth criticality (\ref{fig:SE2_ground_truth}) and network outputs $h_\theta(x,y)$ (\ref{fig:SE2_learned}) are visualized as robot miniatures with color/size proportional to log-criticality (redder and larger is more critical). Critical PRM displays a half order of magnitude improvement in computation time for a fixed success rate (\ref{fig:SE2_time_vs_success}).}
    \label{fig:SE2}
\end{figure}

\emph{SE(2) {\bf L} Problem.} To further demonstrate the ability of the Critical PRM learning methodology to identify narrow passages in the state space from obstacle representations given in the workspace, we also consider the problem of maneuvering a rigid {\bf L}-shaped robot through a constrained environment as depicted in Fig.~\ref{fig:SE2_solution}. The state space $\text{SE}(2) = \mathbb{R}^2\times\mathbb{S}^1$ includes an orientation dimension in addition to two position dimensions. To highlight the complications that arise from considering orientation, the family of environments for this subsection is constructed by computing Voronoi diagrams for randomly drawn points in the unit square and cutting passages in the borders between regions (which have variable orientation by construction) significantly narrower than the lengths the robot body segments.

From Fig.~\ref{fig:SE2_learned}, which displays inferred sample criticality superimposed on a PRM not within the neural net's training set, we can see that the learning process has gained the insight that states where the arms of the {\bf L} straddle a passage are most valuable. The learned model produces smoother criticality predictions compared to the ground truth values depicted in Fig.~\ref{fig:SE2_ground_truth}. This ground truth is in a sense overfit to the specific sample set and associated graph (see, e.g., narrow passages with nearby states having near-zero criticality because the ground truth graph contains no path through the passage); in contrast the regression estimate gives a sense of how a sample might be useful in a generic PRM.
For this problem family we find that putting learned criticality to work in Critical PRM improves the required computation time for a given success rate by approximately half an order of magnitude, Fig.~\ref{fig:SE2_time_vs_success}.

\begin{figure}[t]
    \centering
    \begin{subfigure}[b]{0.2\textwidth}
        \includegraphics[width=\textwidth]{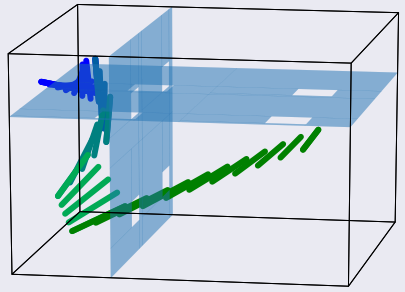}
        \caption{SE(3)}
        \label{fig:se3_problem}
    \end{subfigure}
    $\qquad\quad$
    \begin{subfigure}[b]{0.141\textwidth}
        \includegraphics[width=\textwidth]{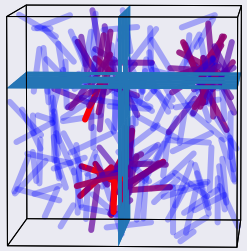}
        \caption{Learned Crit.}
        \label{fig:se3_crit}
    \end{subfigure}
    \begin{subfigure}[b]{0.235\textwidth}
        \includegraphics[width=\textwidth]{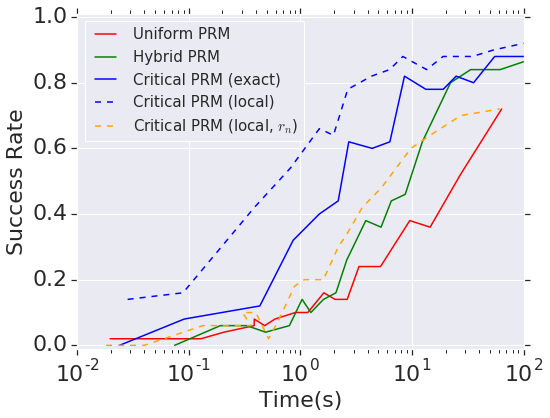}
        \caption{Time (s) vs. Success}
        \label{fig:se3_time_vs_success}
    \end{subfigure}
    \begin{subfigure}[b]{0.235\textwidth}
        \includegraphics[width=\textwidth]{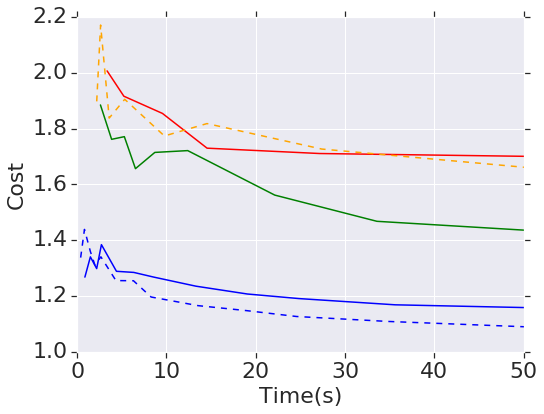}
        \caption{Time (s) vs. Cost}
        \label{fig:se3_time_vs_cost}
    \end{subfigure}
    \caption{
    This SE(3) {\bf I} rigid body planning problem demonstrates that Critical PRM can consider the full state space when selecting samples and again demonstrates the benefit of both critical samples and the hierarchical roadmap. In the learned criticality visualization (\ref{fig:se3_crit}) redder is more critical.}
    \label{fig:se3}
\end{figure}

\emph{SE(3) {\bf I} Problem.} We also consider an $SE(3) = \mathbb{R}^3\times\mathbb{S}^3$ {\bf I} rigid body shown in Fig. \ref{fig:se3}.
We use the same environment inputs and network architecture from the earlier 3D planning problem.
Due to the complexity gathering data for the problem, we only train the criticality network from 200 environments for 100k total samples (and to showcase the minimal data requirements).
Fig. \ref{fig:se3_crit} shows several critical states for the SE(3) problem. The non-critical samples generally occur in free space (blue) and the critical samples (red) tend to be near the gaps and oriented perpendicular to the obstacle plane, though they are reasonably varied, allowing some diversity.
The results are shown in Figs. \ref{fig:se3_time_vs_success}-\ref{fig:se3_time_vs_cost}.
For this environment, the local representation again outperforms by orders of magnitude (even in such a low data regime and with a high dimensional input)--the difference is particularly stark in path cost compared to Hybrid PRM and Uniform PRM.
The global connections too have a large effect in both success rate and cost.

\subsection{Reinforcement Learned Local Policy}\label{sec:prmrl}

\begin{figure}[t]
    \includegraphics[width=0.49\textwidth]{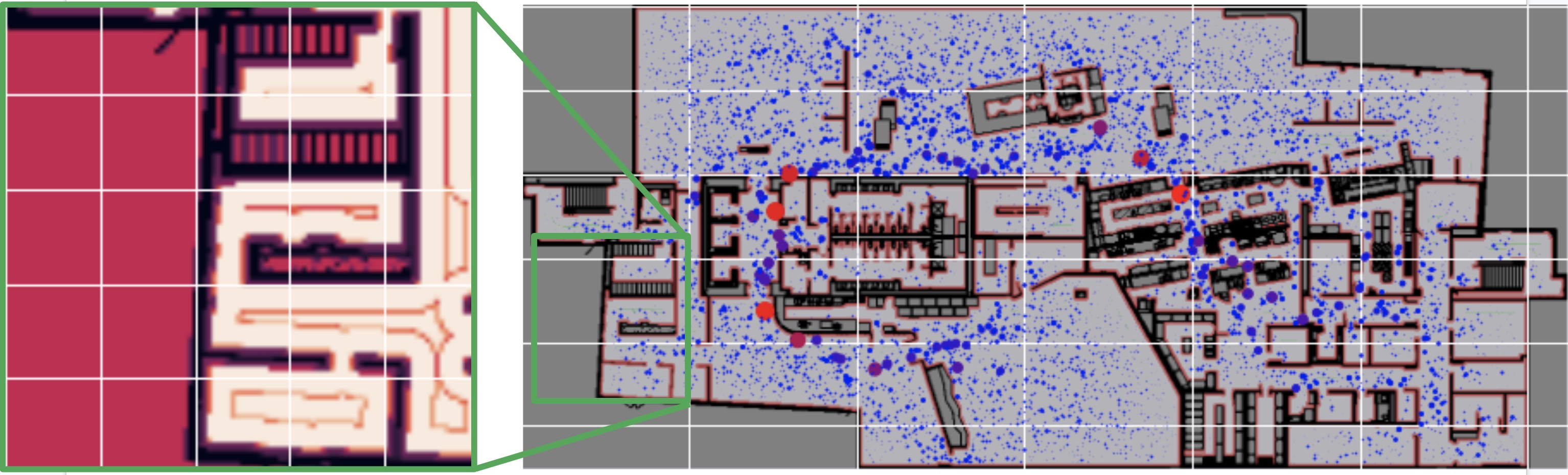}
    \caption{\small 
    Floor plan environment and critical sample data for RL-trained policy. The input to the criticality prediction network is a 100$\times$100 pixel (10m$\times$10m) image of the local environment around a sample. Note the most critical samples avoid open regions and are not necessarily at doorways due to the local intelligent policy.}
    \label{fig:prmrl_train}
\end{figure}

In this section, we demonstrate the generality and efficacy of Critical PRMs on a challenging planning problem that requires consideration of: (1) the effect of a complex local planner, (2) the effect of uncertainty in choosing robust samples that are navigable in the presence of sensor and dynamics noise, and (3) high-dimensional, complex, real-world environment representations in the form of images of office building floor plans (Fig. \ref{fig:prmrl_train}).
This problem follows the formulation presented in \cite{faust2018prm,francis2019long}, in which a differential drive robot navigates an office environment to a local goal via a reinforcement-learned policy.
The policy takes as input a vector from the current state to the goal and a 64-wide, 220$^\circ$ field-of-view to see the local environment along with a realistic noise model.
This is incorporated into the PRM framework by connecting local samples, thus allowing more intelligent local planning. 
These edges are added if the local planner is able to perform the connection 100\% of the time over 20 trials, thus requiring samples to be chosen robustly to the local policy and noise.
The value of $\lambda$ was set as 15 (note the increase due to the complexity of the problem and number of narrow passages).
The critical training data is visualized in Fig. \ref{fig:prmrl_train} along with the input to the network (a 100$\times$100 pixel image over a 10m$\times$10m area). 
The network used on this data trained over three different office floor plans, with a total of 50k samples.
We note that due to the scarcity of environments and complexity of input for the full environments, previous learned methods \cite{ichter2018learning,kumar2019lego} that require full environment input cannot generalize.
Furthermore, approaches that seek to find samples near obstacles or narrow passages \cite{hsu2005hybrid} are likely to both select difficult to reach states given the problem's stochasticity and not extract the key features of the problem given the clutter. 

\begin{figure}[t]
    \centering
    \begin{subfigure}[b]{0.093\textwidth}
        \includegraphics[width=\textwidth]{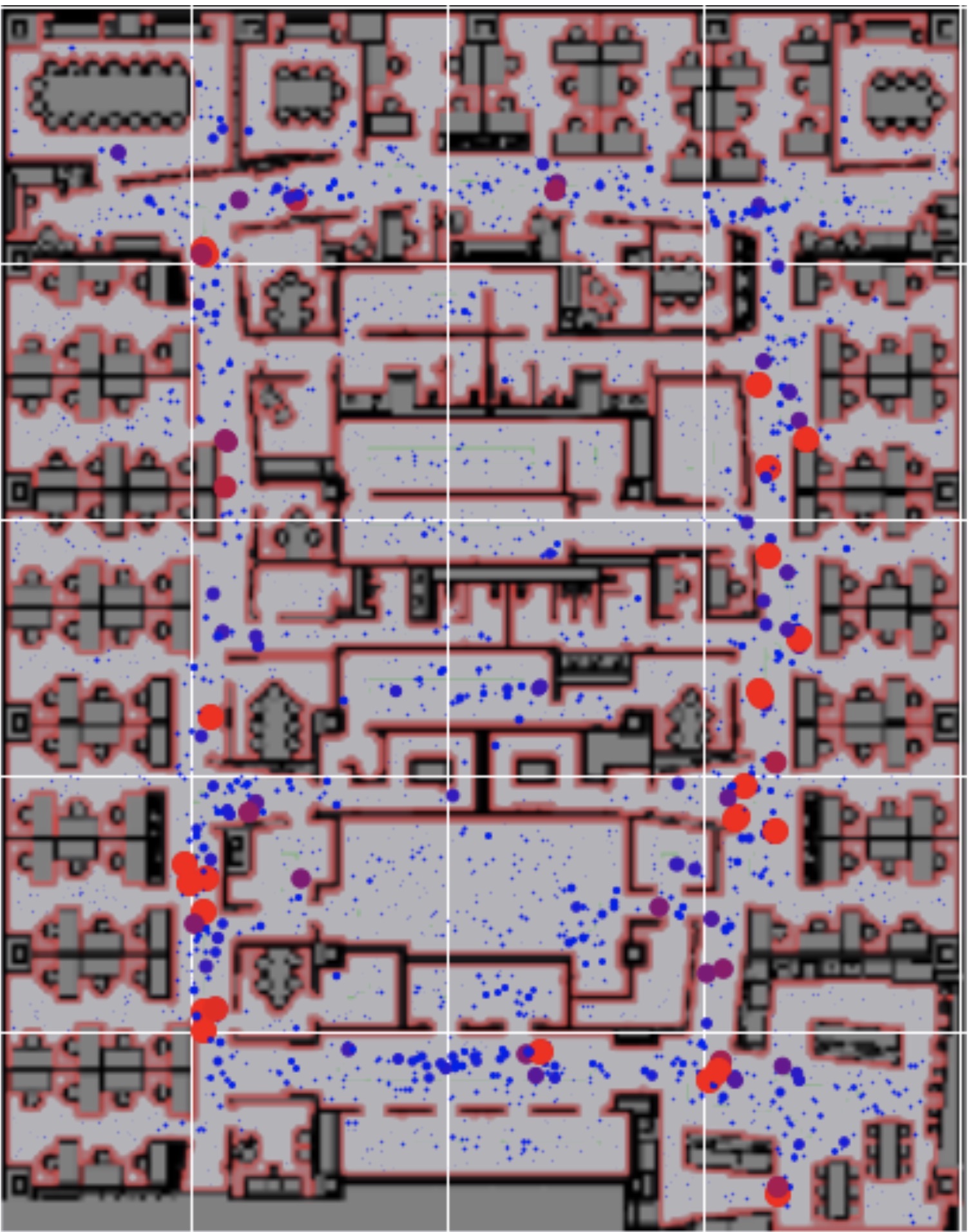}
        \caption{$h_\theta$}
        \label{fig:prmrl_test}
    \end{subfigure}
    \begin{subfigure}[b]{0.189\textwidth}
        \includegraphics[width=\textwidth]{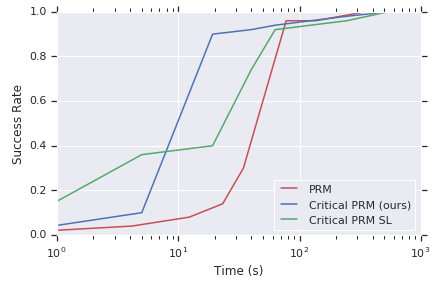}
        \caption{Time vs. Success}
        \label{fig:sfo_prm_rl_time_success}
    \end{subfigure}
    \begin{subfigure}[b]{0.189\textwidth}
        \includegraphics[width=\textwidth]{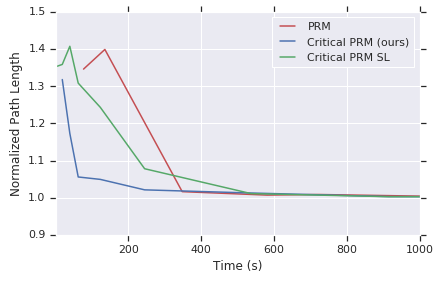}
        \caption{Time vs. Cost}
        \label{fig:sfo_prm_rl_time_cost}
    \end{subfigure}
    \caption{ 
    (\ref{fig:prmrl_test}) Predicted sample criticality for a new floor plan extracts the importance of the center hallway and avoids outside cubicles.
    (\ref{fig:sfo_prm_rl_time_success}--\ref{fig:sfo_prm_rl_time_cost}) Success rate and cost versus time (s) over 50 problems.
    }
    \label{fig:prmrl_results}
\end{figure}

Fig. \ref{fig:prmrl_results} shows the results of Critical PRM on a new building. 
Though the building has not been seen before, the neural network is able to extract the importance of features like hallways and ignore areas around cubicles.
Note that the critical points are not always in doorways as one may expect for a straight line policy, as the intelligent local policy is often able to robustly enter or exit such narrow passages.
In terms of success rate and cost, Critical PRM is approximately 5 times faster to compute.
We also compare to if the critical points were learned only considering straight line (SL) connections (though at runtime, the robot still executes the learned policy).
This compares a method that cannot take the local planner into account when learning to sample.
The straight line connections still outperform PRM, but are approximately 3 times slower than Critical PRM.

\subsection{Physical Experiments}\label{sec:phys}

\begin{figure}[h]
    \centering
    \begin{subfigure}{0.135\textwidth}
        \includegraphics[width=\textwidth]{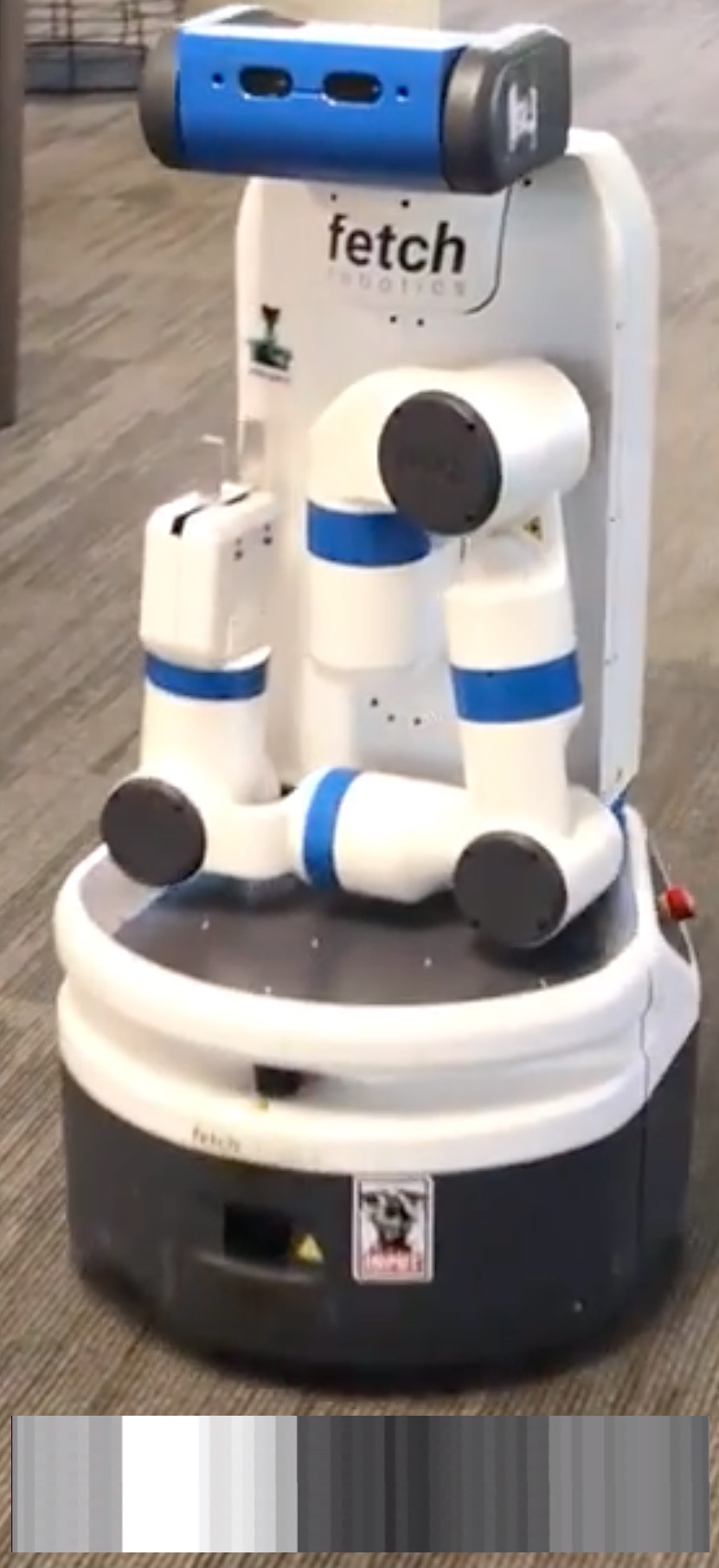}
        \caption{}
        \label{fig:fetch}
    \end{subfigure}
    \begin{subfigure}{0.142\textwidth}
        \includegraphics[width=\textwidth]{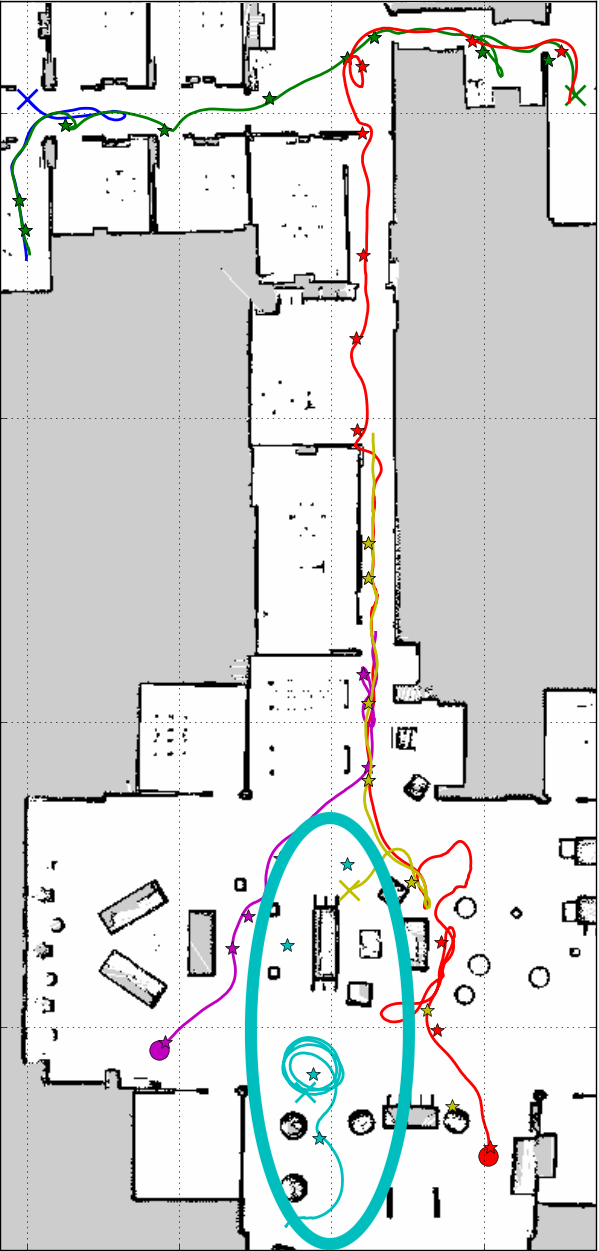}
        \caption{}
        \label{fig:exp_uniform_all}
    \end{subfigure}
    \begin{subfigure}{0.142\textwidth}
        \includegraphics[width=\textwidth]{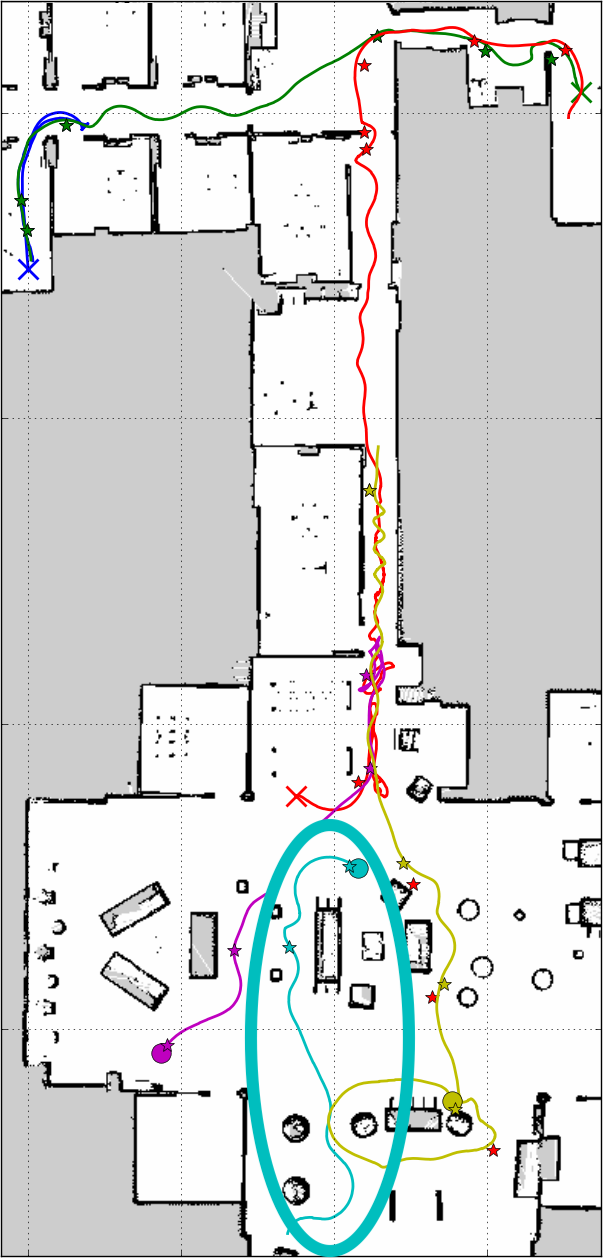}
        \caption{}
        \label{fig:exp_critical_all}
    \end{subfigure}
    \caption{ 
    On robot experiments with Critical PRM computes lower-cost, more-robust trajectories in less time than PRM.
    (\ref{fig:fetch})  Fetch and lidar observation. 
    (\ref{fig:exp_uniform_all}) PRM  and (\ref{fig:exp_critical_all}) Critical PRM trajectories (planned waypoints shown as stars).
    The circled cyan trajectories were run 5 times each, resulting in 40\% success for uniform PRM and 100\% for Critical PRM.
    }
    \label{fig:exp}
\end{figure}

Lastly, we implement the approach on a physical robot, a Fetch operating in an indoor environment using SLAM (see Fig. \ref{fig:exp}).
The robot observes and localizes against the environment with a 220$^\circ$ field-of-view single plane lidar observation and navigates locally using the reinforcement learned policy described in Section \ref{sec:prmrl}.
The input to the criticality network is a 80$\times$80 pixel image of the environment.
The network was trained on 5k samples from a different lidar-mapped environment.

The experiments are shown in Fig. \ref{fig:exp}, for which 6 long-range trajectories are shown traversing the space.
The Critical PRM trajectories in Fig. \ref{fig:exp_critical_all} were generally more direct and robust than their uniform PRM counterpart in Fig. \ref{fig:exp_uniform_all}.
To this end, the cyan (circled) path in Figs. \ref{fig:exp_critical_all}-\ref{fig:exp_uniform_all} was repeated five times, of which the Critical PRM trajectory was successful 100\% of the time versus 40\% for the uniform PRM. 
This improved performance was primarily a result of the Critical PRM using waypoints only when necessary, allowing more spread out waypoints (e.g., when the environment was less cluttered), allowing the RL policy to better adjust to stochasticity in operation.

\section{Conclusions and Future Work}\label{sec:conclusions}

\textit{Conclusions.}
In this paper, we have presented a method towards learning critical samples for sampling-based robot motion planning and using them more heavily.
Specifically, we identify critical samples via betweenness centrality---a graph theoretic measure of a node's importance to shortest paths---and learn to predict them via a neural network that takes local workspace information as input.
For a new planning problem, these critical samples are then sampled from more frequently and connected globally (along with a bed of uniform samples locally connected).
The result is a hierarchical roadmap we term the Critical PRM.
The algorithm is demonstrated to achieve up to a three order of magnitude improvement in computation time required to achieve a success rate and cost due to both the selection of better samples and the use within a hierarchical roadmap.
The method is further demonstrated to be general enough to handle real-world data, state space sampling, and complex local policies.
Lastly, Critical PRM is shown on robot to compute lower-cost and more-robust trajectories.

\textit{Future Work.} 
In the future we plan to extend this work in several ways.
First, we plan to show its ability to identify samples for differentially constrained systems and robot arms.
Second, we plan to investigate how critical samples may be clustered to identify critical regions and reduce overlap between samples.
Finally, we plan to use these results within a hierarchical RL framework by biasing the high level policy towards more critical subgoals.

\bibliographystyle{IEEEtran-short}
\bibliography{cite}

\section*{Acknowledgements}
The authors thank Vincent Vanhoucke, Chase Kew, James Harrison, and Marco Pavone for insightful discussions.

\newpage
\section*{Appendix}

The following describes the data used, algorithm hyperparameters, and network architectures. A dropout layer set to 0.1 follows each layer. Each layer other than the output layer or max-pooling has ReLU activation functions.

\noindent\emph{2D narrow passage, Section \ref{subsec:NPE}:}
\begin{itemize}
\item Input Exact: $(x,y)$-coordinates of each gap
\item Input Local: $10^2$ local occupancy grid from $100^2$
\item Architecture: 2048 - 1024 - 512 - 1
\item Data: 1k environments, 100k samples
\item Alg. Params.: $\Gamma=10$, $\lambda = 2$
\end{itemize}

\noindent\emph{3D narrow passage, Section \ref{subsec:NPE}:}
\begin{itemize}
\item Input Exact: $(x,y,z)$-coordinates of each gap
\item Input Local: $12^3$ local occupancy grid from $36^3$
\item Architecture Exact: 512 - 512 - 512 - 512 - 1
\item Architecture Local: $16\times4^3$ conv - $16\times4^3$ conv - 128 - 128 - 1
\item Data: 1k environments, 100k samples
\item Alg. Params.: $\Gamma=10$, $\lambda = 10$
\end{itemize}

\noindent\emph{SE(2), Section \ref{sec:rigid}:}
\begin{itemize}
\item Input: $33^2$ local occupancy grid from $100^2$ and local position within grid cell, orientation of each sample
\item Architecture: $16\times3^2$ conv - $2^2$ max-pooling - $32\times3^2$ conv - $2^2$ max-pooling - $32\times3^2$ conv - $2^2$ max-pooling - 1
\item Data: 1k environments, 6m samples
\item Alg. Params.: $\Gamma=10$, $\lambda = 5$
\end{itemize}

\noindent\emph{SE(3), Section \ref{sec:rigid}:}
\begin{itemize}
\item Input Exact: $(x,y,z)$-coordinates of each gap and orientation of each sample
\item Input Local: $12^3$ local occupancy grid from $36^3$ and orientation of each sample
\item Architecture Exact: 512 - 512 - 512 - 512 - 1
\item Architecture Local: 
\begin{itemize}
    \item Environment Network [$16\times4^3$ conv - $16\times4^3$ conv - $16\times4^3$ conv]
    \item Orientation Network [256 - 256]
    \item Stack output of networks into [256 - 256 - 1]
\end{itemize}
\item Data: 200 environments, 100k samples
\item Alg. Params.: $\Gamma=10$, $\lambda = 10$
\end{itemize}

\noindent\emph{PRM-RL, Section \ref{sec:prmrl}:}
\begin{itemize}
\item Input: $100^2$ pixel local image ($10^2$ m)
\item Architecture: $16\times8^2$ conv - $2^2$ max-pooling - $32\times8^2$ conv - $32\times8^2$ conv - 128 - 128 - 1
\item Data: 3 environments, 50k samples
\item Alg. Params.: $\Gamma=10$, $\lambda = 15$
\end{itemize}

\noindent\emph{Physical Experiments, Section \ref{sec:phys}:}
\begin{itemize}
\item Input: $80^2$ pixel local image ($8^2$ m)
\item Architecture: $16\times8^2$ conv - $2^2$ max-pooling - $32\times8^2$ conv - $32\times8^2$ conv - 128 - 128 - 1
\item Data: 2 environments, 5k samples
\item Alg. Params.: $\Gamma=10$, $\lambda = 15$
\end{itemize}

\end{document}